\documentclass{article}

\usepackage{arxiv}

\usepackage{latexsym}
\usepackage{amssymb}
\usepackage{amsmath}
\usepackage{amsthm}
\usepackage{booktabs}
\usepackage{enumitem}
\usepackage{graphicx}
\usepackage{color}
\usepackage{times}
\usepackage{soul}
\usepackage{url}
\usepackage[hidelinks]{hyperref}
\usepackage[utf8]{inputenc}
\usepackage[small]{caption}
\usepackage{graphicx}
\usepackage{amssymb}
\usepackage{amsmath}
\usepackage{amsthm}
\usepackage{booktabs}
\usepackage[ruled,linesnumbered,noend]{algorithm2e}
\usepackage{algorithmic}
\usepackage[switch]{lineno}
\usepackage{natbib}

\newcommand\commentout[1]{}
\newcommand\eff{\mathit{eff}}
\newcommand\hff{h_{\mathit{ff}}}
\newcommand\hadd{h_{\mathit{add}}}
\newcommand\hmax{h_{\mathit{max}}}
\newcommand\qmdp{Q_{\mathit{MDP}}}

\title{Heuristics for Partially Observable Stochastic Contingent Planning}

\author{
Guy Shani\\
eBay Research \& \\
Computer and Information Sciences\\
Ben Gurion University\\ 
\texttt{shanigu@bgu.ac.il, gshani@ebay.com.}
}

\begin{document}
\maketitle
\begin{abstract}
Acting to complete tasks in stochastic partially observable domains is an important problem in artificial intelligence, and is often formulated as a goal-based POMDP. Goal-based POMDPs can be solved using the RTDP-BEL algorithm, that operates by running forward trajectories from the initial belief to the goal. These trajectories can be guided by a heuristic, and more accurate heuristics can result in significantly faster convergence.
In this paper, we develop a heuristic function that leverages the structured representation of domain models.  We compute, in a relaxed space, a plan to achieve the goal, while taking into account the value of information, as well as the stochastic effects. We provide experiments showing that while our heuristic is slower to compute, it requires an order of magnitude less trajectories before convergence. Overall, it thus speeds up RTDP-BEL, particularly in problems where significant information gathering is needed.
\end{abstract}

\section{Introduction}

Autonomous agents acting to achieve their goals in stochastic environments often face partial observability, where important information for completing the task is hidden, but can be sensed. The agent must consider the value of information, that is, whether it is better, in terms of expected cost, to invest effort in sensing for the hidden information, or to try and reach the goal without the information.

Consider, for example, a robot navigating in an uncertain environment, where the ground may be unstable. The robot can either operate a sensor that  examines the ground to check its stability, later allowing it to move rapidly only over stable areas, or move slowly and carefully without operating the sensor first. The choice as to whether it is cost effective to use the sensor depends on the probability of unstable areas, the time it takes for the sensor to scan the ground, and the cost of moving through possibly unstable areas. This tradeoff is known as the value of information.

Partially observable Markov decision processes (POMDPs) \citep{sondik1971optimal} are a mathematical model designed for properly estimating the value of information over long planning horizons in such environments. Goal-POMDPs \citep{geffner1998solving} are a subclass where the agent must achieve a goal, upon which the execution terminates. POMDPs can be either specified using a "flat" representation, where each state is a unique entity, or using a factored representation, where a state is composed of state variables. We focus here on a particular factored representation using an extension of the PDDL (Planning Domain Description Language) representation which is popular in the automated planning literature \citep{haslum2019introduction}. PDDL defines actions by preconditions and effects, allowing for a compact representation of complex problems. 

POMDP solvers often use the concept of a belief state, a distribution over the currently possible world states, and compute a value function, mapping a belief state $b$ to the expected cost to the goal from that belief. The RTDP-BEL (Real-Time Dynamic Programming in Belief Space) algorithm \citep{bonet2009solving} computes a value function by running trajectories in belief space from the initial belief to a goal. It depends on a heuristic to guide the trajectories towards goal states. Good trajectories that visit belief states that are typically visited on an optimal path to the goal can lead to much faster convergence. We develop here heuristics that steer RTDP-BEL towards such good trajectories.

An advantage of using a PDDL representation is that it paves the way to employing the multitude of heuristics developed for various automated planning solvers. For example, the delete relaxation approach \citep{BlumF97}, that ignores negative effects, can be used to rapidly estimate the distance to the goal using techniques such as $\hmax$ \citep{bonet2001planning} and $\hff$ \citep{hoffmann2001ff}. This paper suggests how these techniques can be used for generating heuristic cost-to-go estimations for belief states.

More specifically, during the heuristic computation, we show how in the forward exploration in relaxed space we can take sensing actions into account, considering only actions whose preconditions hold in all possible states. Thus, we can compute an heuristic estimation of the information needed for achieving a particular plan. This results in more accurate cost-to-go heuristic estimations.

We suggest here heuristics and test them in the context of the RDTP-BEL offline POMDP solver, which has shown superior performance compared to the popular point-based approaches in goal-based domains \citep{bonet2009solving}. We hence do not evaluate our methods here in the context of point-based solvers.
Although online POMDP solvers \citep{silver2010monte} have gained much popularity, offline planning has a number of advantages, mainly low online computational burden, allowing the previously computed policy to be rapidly employed in weak end devices.
Our empirical evaluation compares our heuristics to popular methods, such as the well known $\qmdp$ heuristic \citep{bonet2000planning,hauskrecht2000value}, and focusing only on the most likely state. We show that our heuristics, while slower to compute, improves the convergence of RTDP-BEL such that the overall runtime is significantly reduced, especially in tasks that require costly information gathering action sequences.

\section{Background}

A goal Markov decision process (goal-MDP) \citep{bellman1957markovian,bertsekas2012dynamic} is a tuple $\langle S, S_0,A,tr,G,C \rangle$ where $S$ is a state space, $A$ is an action, $tr:S\times A \times S \rightarrow [0,1]$ is a transition function. $tr(s,a,s')$ is the probability of executing action $a$ at state $s$ and reaching state $s'$. $S_0\subset S$ is a set of possible initial states, and $G \subset S$ is a set of goal states, that the agent must reach. $C:S \times A \rightarrow \mathbb{R}$ is a cost function. $C(s,a)$ is the cost for applying action $a$ at state $s$. In goal-MDPs for $s\in G$, $C(s,a)=0$ and $C(s,a)>0$ otherwise.

A solution to a goal MDP is a policy $\pi:S \rightarrow A$ that assigns an action to every state. One can create a policy by computing a value function $V:S\rightarrow \mathbb{R}$ assigning a value for each state. This value function is optimal if $V(s)$ is the minimal expected cost of reaching a goal state from $s$. The Bellman backup:
\begin{equation}
    V(s) \leftarrow \min_{a\in A} C(s,a) + \sum_{s' \in S} tr(s,a,s') V(s')
\end{equation}
can be used to iteratively update the value function. Value iteration applies the Bellman update over all states repeatedly, and converges to the optimal value function.

Real Time Dynamic Programming (RTDP) \citep{barto1995learning} applies the Bellman update only along trajectories that start at a state in $S_0$ and terminate at a state in $G$. Thus, RTDP is able to focus on states that are visited along an optimal path to the goal, avoiding repeated Bellman backups over other states, that are less important. This allows RTDP to converge much faster than value iteration in many domains.

A goal-POMDP is a tuple $\langle S, S_0,b_0,A,tr,G,C,\Omega, O \rangle$ where $\langle S, S_0,A,tr,G,C \rangle$ is a goal-MDP, known as the underlying MDP. $b_0$ is a distribution over the possible initial states in $S_0$, $\Omega$ is a set of possible observations, and $O(a,s,o)$ is the probability of observing $o$ after applying action $a$ and reaching state $s$. In this paper we focus on deterministic observations, and hence, $O(a,s,o)\in[0,1]$. This is not truly a limitation, because every POMDP with stochastic observations can be translated into a deterministic observation POMDP, by adding a state variable whose value changes stochastically, yet observed deterministically. Consider for example a noisy wall detection sensor. Instead of observing a wall with stochastic errors, we can add a red light to the sensor, which serves as the state variable. The red light is stochastically lit by the sensor when there the sensor detects a wall, but the agent observes whether the light is on deterministically.

A belief state is a distribution over states, where $b(s)$ is the probability of state $s$. Given a belief $b$ an action $a$ and an observation $o$ one can compute the new belief state $b^o_a$ using:
\begin{equation}
    b^o_a(s') = 
    \begin{cases}
            \kappa  \cdot \sum_s b(s)tr(s,a,s') : O(a,s',o)=1\\
            0 : O(a,s',o)=0
    \end{cases}
\end{equation}
where $\kappa$ is a normalizing factor.
A popular technique for solving POMDPs is by computing a value function over the belief space MDP, defined over the all belief states. While this MDP has an infinite amount of states, its value function is convex, which allows for a convenient representation using a set of half-spaces called $\alpha$-vectors.

\begin{algorithm}[t]
\caption{RTDP-BEL}
    \label{alg:RTDP-BEL}
\Indp\Indpp
\SetKwBlock{RTDP}{RTDP-BEL}{end}
\RTDP{
    \While{$V$ has not converged}{
        $b \leftarrow b_0$\\
        $s \leftarrow$ sample from $b_0$\\
        $l \leftarrow $ the empty list\\
        \While{$s \notin G$}{
            add $b$ to $l$\\
            $a^* \leftarrow argmin_a{ C(b,a)+  \sum_o pr(o|a,b)V(b^o_a)} $ \\
            $V(b) \leftarrow C(b,a^*)+ \sum_{o}{pr(o|b,a^*)V(b^o_{a^*})}$\\
            $s' \leftarrow $ sample from $tr(s, a^{*}, \cdot)$\\
            $o \leftarrow $ the observation s.t. $O(a,s',o)=1$\\
            $b \leftarrow b^0_a$
        }
        \ForEach{$b \in l$ in reversed order}{
            \ForEach{$a \in A$}{
                $a^* \leftarrow argmin_a{ \sum_o pr(o|a,b)V(b^o_a)} $ \\
                $V(b) \leftarrow C(b,a^*)+ \sum_{o}{pr(o|b,a^*)V(b^o_{a^*})}$\\
            }
        }
    }
}
\Indp\Indpp
\end{algorithm}

In this paper, however, we do not use $\alpha$-vectors. Instead, we use RTDP in belief space, called RTDP-BEL \citep{geffner1998solving,bonet2000planning,bonet2009solving} to focus the exploration of the belief space to beliefs that are likely to occur along an optimal path to the goal. RTDP-BEL is presented in algorithm~\ref{alg:RTDP-BEL}. Lines 6-12 produce a forward trajectory. The values along that trajectory are updated when moving forward, and thus, one can show that the trajectory must terminate at a goal state. It is useful, however, to also update the states after the trajectory has terminated moving backward from the goal (lines 13-5).

In the value function update (lines 9,15) when $V(b)$ was not yet computed, we can use a cost-to-go heuristic, to initialize it. If this heuristic is well correlated with the optimal value of $V(b)$, it can steer RTDP-BEL to chose good actions in line 8 that lead it faster to the goal. The heuristic must be optimal (a lower bound) over the optimal costs, otherwise, when an action may seem optimal, RTDP-BEL will stop exploring alternative actions from that belief, and may converge to a sub-optimal value function. As we shall later see, this is a real problem for some heuristics. Our paper focuses on developing strong heuristics for initializing $V(b)$.

Given a value function $V:B\rightarrow \mathbb{R}$ computed by RTDP-BEL, we can define a policy using:
\begin{equation}
\label{eqn:policy}
    \pi_V(b) = argmin_a C(b,a) + \sum_o pr(o|b,a) V(b^o_a)
\end{equation}
Unlike value functions represented using $\alpha$-vectors, a value function represented using a direct mapping from belief states to values, does not generalize to other belief states. Thus, when computing the policy for a belief $b$, it might be that some values $V(b^o_a)$ are missing. In this case, we can use the heuristic again for these missing values.

\commentout{
\begin{algorithm}[ht]
\caption{RTDP}
    \label{alg:RTDP}
\footnotesize

\SetKwBlock{RTDP}{RTDP}{end}
\RTDP{
    Init $Q(s,a)$ to an upper bound \\
    $V(s) \leftarrow \max_a{Q(s,a)}$\\
    \While{$V$ has not converged}{
        $s \leftarrow s_0$ \\
        \While{$s$ is not terminal state}{
            $a^* \leftarrow argmax_a{Q(s,a)} $ \\
            $s' \leftarrow $ sample from $tr(s, a^{*}, \cdot)$\\
            $Q(s,a^*) \leftarrow R(s,a^*)+ \gamma \sum_{s'}{tr(s,a,s')V(s')}$\\
            $V(s) \leftarrow \max_a{Q(s,a)}$\\
            $s'\leftarrow$ sample from $tr(s,a,\cdot)$\\
            $s \leftarrow s'$
        }    
    }
}

\end{algorithm}
}

POMDPs can also be described in a structured manner, following conventions developed in the planning community, specifically for stochastic planning and contingent planning with sensing actions \citep{albore2009translation}. We now provide such a specification. A stochastic, partially observable, contingent planning problem is a tuple (overloading previous definitions) $\langle P,A,G,I \rangle$. $P$ is a set of facts. Each state is a truth assignment for all facts.

$A$ is a set of actions. Each action $a\in A$ is a tuple $\langle pre,\eff,obs \rangle$ where $pre$ is the precondition, a set of facts that must hold before applying the action. $\eff$ is a set of conditional effects of the form $(c,e)$ where $c$ is a set of facts that form the condition --- if all facts in $c$ hold then the condition is fulfilled. $e$ is a probabilistic formula --- a set $(e_1,...,e_k)$ of possible probabilistic effects, where $e_i$ is a set of facts.
For each fact $l$ in $e_i$, if $l$ was false prior to the execution of the action, it would become true with probability $pr(e_i)$ after the action is executed.
$obs$ is a set of facts whose value is observed in the new state following the action. Action have a cost $C(a)$, $C(s,a)=C(a), s \notin G$, and $C(s,a)=0, s \in G$.

We assume here, as is often done in automated planning literature, that all preconditions and conditions contain only positive facts. This is not truly a limitation, as one can always maintain for every fact $p$ an auxiliary fact $np$, ensuring that $p=\neg np$ always holds.

$I$ is a a set of probabilistic formulas over the facts, defining the possible initial values of facts, thus also defining the initial belief state. $G$ is a set of goal facts.

The agent operates in belief space. Hence, before applying an action $a$ in a belief $b$, the agent must ensure that the preconditions $a.pre$ hold in every state $s$ such that $b(s)>0$. This approach is useful for limiting the branching factor --- the amount of actions that must be considered at every belief state. 
As we assume here deterministic observations, each observation effectively eliminates from the next belief every state that does not agree with the observed value of one of the facts in $a.obs$.

\section{Related Work}

In this paper we use our heuristics within the RTDP-BEL algorithm. A popular alternative to RTDP-BEL is the point-based approach, where a value function in the form of a set of $\alpha$-vectors is computed over a finite set of belief states \citep{shani2013survey,SmithS04,pineau2006anytime,kurniawati2008sarsop,spaan2005perseus,poupart2011closing}. While RTDP-BEL is well suited for goal-POMDPs, point-based methods may also work well in such domains.
It is quite possible that our heuristics can be helpful in obtaining good belief states for point-based approaches as well. That is, one may be able to design a new point-based algorithm that uses our methods as a heuristic for collecting belief points. We leave this for future research. 

Another approach for acting in partially observable environments is through online approaches, such as POMCP \citep{silver2010monte} and DESPOT \citep{somani2013despot}. Online algorithms expand a belief tree from the current belief state, and then choose an action. When constructing the tree, heuristics are very useful. However, in many of the domains that we experiment with, the number of possible actions at each belief is huge. This renders online algorithms practically useless, because with a huge branching factor, the trees grow beyond manageable sizes. Attempts at using such heuristics in POMCP over similar domains did not scale beyond tiny problems \citep{blumenthal2023rollout,blumenthal2024rollout}.

\citet{bertoli2002improving} consider similar motivations to our approach when developing heuristics for search in belief space. They capture knowledge at a higher degree than the state based level, estimating the probability that some formulas hold. They estimate a value for sensing actions only at a myopic level, which RTDP-BEL already handles.

\citet{bryce2004planning,bryce2006planning} extend methods for computing heuristics in classical planning to belief space. They use the planning graph structure, which is a more elaborate construct than the simple layered graph used for computing $\hff$. To adapt the planning graph to belief space, they show how to compute a heuristic estimate for multiple states together. In this context, however, sensing action provide no value.
We evaluate this idea, although not the efficient computation, in our experiments below, showing that it does not handle well domains requiring information gathering.

\citet{WangD11} also provide a myopic value of information estimation, estimating the long range effects of an immediate sensing action. They do not, however, take into account sensing actions that will be needed later in the plan.

Our belief maintenance method is simple, maintaining the distribution over all states that have a non-zero probability for each belief. In many domains, the number of non-zero probability states is not very large, so this representation is not very costly. We are also caching state transitions once they were computed to avoid repeated computations.
Our attempts at using more sophisticated methods (e.g. \citep{brafman2016online}) did not show any advantage. In any case, the bottleneck in our algorithms is the heuristic computation, not the belief maintenance.
That being said, many methods were suggested for more efficient representations \citep{BonetG16,to2011effectiveness}, and we leave the examination of such methods in the context of RTDP-BEL for future research.

\citet{kim2019pomhdp} also investigate the use of heuristics in RTDP-BEL. They suggest to use multiple heuristics together, one admissible, and some non admissible, but more accurate. They show that they can still guarantee convergence, while taking into account the more accurate heuristic. In their experiments, they use domain-specific heuristics, while we focus here on domain independent heuristics. One may possibly aggregate some of the heuristics that we evaluate, e.g. our belief-based, state-based, and $\qmdp$ together using their methods.

\citet{saborio2020efficient} use subgoals to construct an heuristic estimate in robotics environments. They compute problem features, and evaluate how many features are satisfied in a given state. They do not, hence, estimate the value of information in sensing actions that require lengthy action sequences to execute from the current belief. Subgoals were investigated in the classical planning literature, e.g., through the computation of landmarks \citep{richter2008landmarks}, but to date have not been able to substantially outperform other heuristics. We leave the investigation of subgoals in the context of RTDP-BEL for future research.

\section{Heuristics In Belief Space}

We now describe our main contribution --- a heuristic in belief space that takes into account the need for sensing actions. Most, if not all, previous heuristics focus on computing a heuristic for each state, and then aggregating these heuristic values. We, however, suggest a method that directly computes a plan in a relaxed belief space. 

Our methods currently assume unit costs. Extensions of our heuristics to non-unit costs is left for future research.

\subsection{State-based Heuristics}
\label{scn:StateHff}

Initializing a heuristic optimistic bound for POMDPs was previously suggested. Perhaps the most popular methods are based on the value function for the underlying MDP \citep{hauskrecht2000value}. Given a computed value function $V_{MDP}$ one can define the heuristic value for a belief state by either focusing on the most likely state: $s_{ML}=argmax_s b(s)$, $h_{ML}(b)=V(s_{ML})$, or the $\qmdp$ heuristic that uses the weighted sum: $h_{QMDP}(b)=\sum_s b(s)V(s)$. More elaborate attempts, such as FIB \citep{hauskrecht2000value}, did not show a substantial improvement over $\qmdp$.

In the planning community, there are many possible heuristics for estimating the cost of reaching the goal. A popular choice is to compute distances in a relaxed problem, known as delete-relaxation \citep{hoffmann2001ff}. In delete-relaxation we ignore negative facts in the effects of actions. Thus, once a fact is added to a state, no action can remove it.

We can hence compute, for a given state, layers of positive facts iteratively (Algorithm~\ref{alg:delete-relaxation}). The first layer $F_0$ contains all the positive facts that hold in the input state $s$ (line 2). Then, we apply all actions that can be executed at $F_0$ (line 7), and their positive effects are added to the next layer $F_1$ (line 8). This is continued until no new facts can be added, or until all goal facts appear in $F_i$ (line 9). Then, we can compute a heuristic estimate. For example, we can return the number of layers until all goal facts were achieved (lines 12-13), known as the $\hmax$ heuristic, or sum the indexes of the layers where each goal fact appeared for the first time, known as the $\hadd$ heuristic \citep{bonet2001planning}. However, $\hadd$ is inadmissible, as it does not acknowledge that several facts can be achieved together. For RTDP, this can be a serious disadvantage, as inadmissible heuristics cause RTDP to converge to a sub optimal solution. 

\begin{algorithm}[t]
\caption{Classical Delete Relaxation Heuristic}
    \label{alg:delete-relaxation}
\SetKwBlock{code}{$\hmax(s)$}{end}
\code{
    $F_0 \leftarrow $ all positive facts in $s$\\
    $i \leftarrow 0$\\
    $done \leftarrow false$\\
    \While{ $\neg done$}{
        $F_{i+1} \leftarrow F_i$\\
        \ForEach{$a \in A$ s.t. $a.pre$ hold in $F_i$}{
            Apply $a$ in $F_i$, add all positive effects to $F_{i+1}$
        }
        \If{$G$ holds in $F_i$ or $F_{i+1}=F_i$}{
            $done \leftarrow true$
        }
        $i \leftarrow i+1$\\
    }
    \If{$G$ holds in $F_i$}{
        \Return $i$
    }
    \Return $\infty$
}
\end{algorithm}

The implementation shown in Algorithm~\ref{alg:delete-relaxation} is simplistic, and there are several ways to improve it. For example, one can add to $F_i$ only new facts that did not appear in some $F_j$, $j<i$. Then, we can check for the next level only actions that have some precondition in $F_j$, because there is no reason to execute an action twice. This is inaccurate for conditional effects that may have been inactive in past layers, and are now becoming active. However, we can take each action with conditional effects and create a set of actions, one for each condition. As in delete relaxation actions are executed in parallel, this is equivalent to executing the original action with multiple conditions activated. Also, as we repeatedly estimate a heuristic for a state $s$ from different belief states, we can cache the state estimations. In many domains, all possible states become cached, and hence, although new beliefs are constantly generated, no new heuristic computations are needed.
We use these, and other, less important, modifications to expedite the computation, but leave them out of the pseudo-code for ease of exposition. 

\citet{bryce2006planning} suggested to extend this method to belief states, by computing, e.g., $\hmax$ for each possible state. Then, a possible heuristic can be the weighted sum of heuristic values: $\hmax(b)=\sum_s b(s)\hmax(s)$. They show how this computation can be done efficiently without handling each state independently, but in our implementation we used a simple approach, computing $\hmax$ for each state and caching results for future computations.

An important disadvantage of $\hmax$ is that it ignores actions that are executed in parallel at the same layer. Thus, it often grossly underestimate the cost to achieving the goal. The $\hff$ heuristic further improves upon $\hmax$ in computing a plan in delete-relaxation space. This plan can be computed by maintaining, for each fact $l$, the action that first generated it. Upon reaching the goal, we trace back, identifying for each goal fact the action that generated it, and then continuing backwards to the precondition facts of that action, until all preconditions are satisfied at the input state. Then, the length of the plan becomes the heuristic estimate for that state. $\hmax$ is admissible in the stochastic sense, i.e. the value it computes is an optimistic estimate of the real expected cost. $\hff$ however is inadmissible, but in the domains that we experiment with, it does not over estimate.

In all the above heuristics, actions that only observe facts, but have no effect, are completely ignored. That is, the heuristic value of observations, and hence, the value of information, cannot be computed. This does not mean, though, that RTDP-BEL ignores these actions. As it iterates over all actions, and over all observations, it computes the myopic value of information of immediate actions. However, these heuristics provide to value to sensing actions that are distant from the current position of the agent.

\subsection{Supporting Stochastic Effects}
\label{sec:stochastic}

Delete-relaxation heuristics were originally designed for deterministic domains, and thus ignore stochastic effects. We suggest the following straight forward method for handling stochastic effects. For a fact $l$ identified at level $i$ that is a stochastic effect of an action with probability $pr(l)$, we insert the fact only at level $F_{i+c}$, where $c=\lceil\frac{1}{pr(l)}\rceil$. That is, we assume that the fact would only be achieved after $c$ executions of the action. This is equivalent to the expected number of times that $a$ must be executed before $l$ is achieved.

When computing our delete relaxation heuristics, we wait until the level where the fact is added, before deciding which action added it. That is, there may be another action that achieved this fact, deterministically or with higher probability, earlier.

\begin{algorithm}[t]
\caption{Belief Delete Relaxation Heuristic}
    \label{alg:SDR}
\footnotesize
\SetKwBlock{bhmax}{$\hmax(b)$}{end}
\bhmax{
    Let $s_{ML}$ be a most likely state in $b$\\
    $Valid=\{s:b(s)>0\}$\\
    \ForEach{$s\in Valid$}{
         $F^s_0 \leftarrow $ all positive facts in $s$\\
    }
    
    $i \leftarrow 0$\\
    $done \leftarrow false$\\
    \While{ $\neg done$}{
        \ForEach{$s\in Valid$}{
            $F^s_{i+1} \leftarrow F^s_i$\\
        }
        \ForEach{$a \in A$ s.t. $a.pre$ hold in $\cap_{s\in Valid} F^s_i$}{
            \ForEach{$s \in Valid$}{
                Apply $a$ in $F^s_i$, add all positive effects to $F^s_{i+1}$      \\
                \If{$a$ is a sensing action}{
                    \If{$\exists p \in a.obs : (p \in F^s_i, p \notin F^{s_{ML}}_i) \vee (p \notin F^s_i, p \in F^{s_{ML}}_i) $}{
                        Remove $s$ from $Valid$\\
                    }
                }

            }
        }
        \If{$G$ holds in $\cap_{s\in Valid} F^s_i$ or $\forall s \in Valid, F^s_{i+1}=F^s_i$}{
            $done \leftarrow true$
        }
        $i \leftarrow i+1$\\
    }
    \If{$G$ holds in $\cap_{s\in Valid} F^s_i$}{
        \Return $i$
    }
    \Return $\infty$
}
\end{algorithm}

\subsection{Belief-based Heuristic}
\label{scn:BeliefHff}

We now describe a new heuristic that is based on the delete-relaxation approach above, but extends it into belief space, allowing us to assign a value for sensing actions. The main difference from the classical method is in maintaining a set of achieved facts for each possible state. An action can be executed only if all states agree that it's preconditions hold. The heuristic computation must reach the goal given all possible states, or identify invalid states --- states that were observed to be inconsistent with the true world state.

We choose one state, e.g. a maximum likelihood state, and compute the heuristic as if it is the real current state. Then, we can eliminate from considerations other states that do not agree with the assumed real state about the value of some observation. As the observations in our model are deterministic, each observation may eliminate some states, and thus may help us in enabling additional actions whose precondition was not met in the eliminated states. This approach allows us to evaluate which sensing actions provide useful information that allows us to identify the true world state.

Algorithm~\ref{alg:SDR} presents the pseudo-code for this heuristic. We begin by choosing a most likely state, denoted $s_{ML}$ to serve as the assumed real world state (line 2). We then define a set $Valid$ initialized to all states with non-zero probability in the current belief state $b$. We maintain for each valid state $s$ the set of positive facts $F^s_i$ (lines 4-5). We then begin computing the layers of positive facts for all states.

We iterate over all actions, whose precondition holds for all currently valid states. That is, we consider only facts that appear in the intersection of all valid $F^s_i$ (line 13). Then, for all valid states, we update $F^s_{i+1}$. Due to conditional actions, different states may acquire different new facts. For ease of exposition, we are not showing in the pseudo-code the treatment of stochastic effects (Section~\ref{sec:stochastic}).

We now consider sensing actions (lines 14-16). For each such action $a$ that observes a set $a.obs$ of facts, we check if there is a fact $p$ with different values in $F^{s_{ML}}_i$ and $F^s_i$ (line 15). If this is the case, then, assuming $s_{ML}$ is the true state of the system, $s$ must be impossible, as observations are deterministic. Thus $s$ is removed from the set $Valid$ of currently valid states (line 16).

This process ensures that, just as the sets $F^s_i$ only grow with additional layers, the set $Valid$ can only become smaller. In many cases $Valid$ eventually consists only of $s_{ML}$ and the rest of the layers are computed in a similar manner to Algorithm~\ref{alg:delete-relaxation}.
We also use the approach suggested above for handling stochastic effects.

For computing a plan in relaxed space for the $\hff$ heuristic, we must again keep track of which action first generated a fact $l$. We consider only facts that hold for all valid states.
Such a new fact $l$ can be introduced in a number of ways; First, an action $a$ can add an effect $l$ to all states, due to unconditional effects. In that case we mark $a$ as the action that generated $l$.

Second, an action can add $l$ to some valid states due to conditional effects. Then, it may be that several actions add $l$ for different valid states until for all valid states $F^s_i$ contains $l$. In that case we mark only the last action that added $l$ as the action that generated $l$. This is imprecise, because it might be that many actions are needed for different valid states. However, this is an optimistic assumption, and thus it cannot cause the heuristic to become non admissible.

Third, it may be that due to an observation, several states have become invalid, and $l$ holds in all remaining valid states. In that case we attribute the achievement of $l$ to the sensing action. Again, this is an optimistic assumption, because it might be that a set of sensing actions, each eliminating some states, was required to ensure that $l$ holds in all current states. As above, such optimistic assumptions cannot cause the heuristic to become non admissible.

We now reconstruct the delete-relaxation plan backwards. Then, we return the number of actions in the plan as the heuristic estimate.

Caching of heuristic computations here is more difficult, but still useful. As $h(b)$ does not depend on the distribution of $b$, we cache the results given $s_{ML}$ and the set of states $\{s:b(s)>0\}$. Hence, if we encounter another belief state that has the same most likely state and the same set of possible states, then we can use the cached heuristic computation. This is much less effective than caching individual state heuristic estimations. However, in stochastic domains, we often encounter beliefs with the same possible states, but with slightly different probabilities. In such cases, caching is very effective.

Our belief delete relaxation heuristics use the same stochastic effects mechanism described earlier as the state-based approaches.

\paragraph{Complexity:}

Computing the delete relaxation graph requires $O(|A|^2)$ operations, assuming a bounded number of action effects, as any action can be executed only once. Hence, computing our heuristics for a belief state $b$ requires $O(|b||A|^2)$ operations, where $|b|$ is the number of states where $b(s)>0$. This is needed for every heuristic estimate computation. Classical heuristics such as $Q_{MDP}$ require solving the underlying MDP, which is in theory polynomial (using linear programming), but given the more practical value iteration requires $O(|S|^2|A|k)$ operations, where $k$ is the number of iterations. Given the value function, computing a heuristic estimation requires $O(|b||A|)$ operations.

\section{Empirical Evaluation}

We now report an empirical evaluation, comparing the performance of the various heuristics that we suggest, in the context of RTDP-BEL. All methods were implemented in C\# within the unified planning framework\footnote{\url{https://github.com/aiplan4eu/up-cpor}, upon acceptance.}.
The experiments were conducted on an $i7$ 3.40GHz CPU with 8 cores, and 64GB RAM.

\paragraph{Methods:}

We compare the following heuristics; First, we use two MDP-based heuristics --- $\qmdp$ and the value of the most likely state (Section~\ref{scn:StateHff}). These heuristics require the computation of a value function for the underlying MDP. In some of our problems, simple value iteration did not converge in under 5 minutes. Using regular RTDP over the state space, on the other hand, is insufficient. This is because RTDP focuses on states that are visited on good paths to the goal in the MDP. However, in the POMDP, there are important states that do not belong to any such path. This is most pronounced in cases where information gathering is needed. We hence use the following compromise --- we compute initially an MDP value function using RTDP, and later, during the execution of RTDP-BEL, when reaching a state that was not visited by the initial RTDP execution, we run RTDP again for 100 additional iterations starting at that state. When reporting runtime, we exclude the initial RTDP execution, as it is not optimized, and other value function methods may be faster.
Below we denote the two MDP based methods as $ML$ for most likely state, and $\qmdp$.

We use two heuristics based on $\hff$ that we describe above --- computing the expected cost over the possible states using $\hff(s)$ (Section~\ref{scn:StateHff}), and computing $\hff(b)$ (Section~\ref{scn:BeliefHff}). We also experimented with $\hmax$ and $\hadd$ for the two cases (state-based and belief-based), but both performed much worse than $\hff$. To avoid overcrowding the table below we do not report these experiments.
We also use a flat uninformative heuristic, that gives a cost of 1 to all non-goal beliefs.

\paragraph{Procedure:}

For each heuristic method we run RTDP-BEL for at most 300 seconds of computation time. We stop the computation repeatedly to evaluate the current policy, to see if the algorithm has converged. The time required for policy evaluation is not considered a part of the runtime of the algorithm. Policy evaluation is run for each possible start state, or for 100 iterations, if there are less than 100 initial states. Each iteration runs from a start state until the current belief satisfies the goal. We stop an iteration if it exceeded 500 steps, assuming a loop.
We select the action $a$ with the best $C(a)+\sum_o pr(o|b,a) V(b^o_a)$, given the current value function $V$. During policy evaluation (Eqn.~\ref{eqn:policy}), if a belief $b$ is encountered for which $V(b)$ was not yet computed, we use the value that the evaluated heuristic assigns to that belief instead.
If 5 consecutive policy evaluations succeeded reaching the goal in all iterations, and the average cost of the policy evaluation has not changed by more than $1\%$, we assume that RTDP-BEL has converged. We then run a policy evaluation over the final value function for 1000 iterations, and compute the average cost, and the number of failed iterations.
We report averages over 50 executions.

\begin{table}[t]
\caption{Problem properties. D/S denotes deterministic / stochastic transitions. U/N denotes uniform / non-uniform initial belief. L denotes localize, EL denotes logistics, CB colorballs, B blocks, D doors, M maze, and W denotes wumpus. $|B|$ estimates the number of beliefs observed during the execution of an optimal policy.}
\label{tbl:domains}
\centering
    \begin{tabular}{|l|c|c|c|c|c|c|}
    \hline

Name&	Type	&$|P|$	&$|S|$	&$|A|$	&$|B|$	&$E[C]$	\\	\hline\hline						
L5	&	DU	&	21	&	70	&	9	&	58	&	12.8	\\	\hline
L5	&	SU	&	21	&	71	&	9	&	69	&	20.01	\\	\hline
L7	&	DU	&	34	&	119	&	9	&	102	&	18.44	\\	\hline
L7	&	SU	&	34	&	120	&	9	&	131	&	22.65	\\	\hline
L9	&	DU	&	46	&	180	&	9	&	164	&	23.97	\\	\hline
L9	&	SU	&	46	&	182	&	9	&	190	&	38.87	\\	\hline
														
EL	&	DU	&	19	&	24,000	&	96	&	1097	&	20.52	\\	\hline
														
CB	&	DU	&	14	&	4240	&	56	&	244	&	15.27	\\	\hline
														
B7	&	DU	&	63	&	37,477	&	504	&	57	&	8.57	\\	\hline
B7	&	DN	&	63	&	37,477	&	504	&	154	&	22.65	\\	\hline
B7	&	SU	&	63	&	37,477	&	504	&	186	&	8.31	\\	\hline
														
D7	&	DU	&	70	&	9457	&	336	&	2532	&	30.89	\\	\hline
D7	&	DN	&	70	&	9457	&	336	&	2597	&	29.27	\\	\hline
D7	&	SN	&	76	&	$8\times10^8$	&	426	&	19	&	20.45	\\	\hline
														
M7,1	&	DU	&	61	&	2368	&	308	&	1526	&	30.2	\\	\hline
M7,2	&	DU	&	63	&	4736	&	308	&	3046	&	32.05	\\	\hline
														
W4	&	DU	&	38	&	540	&	81	&	82	&	12.33	\\	\hline
W4	&	DN	&	38	&	540	&	81	&	94	&	12.22	\\	\hline
W5	&	DU	&	47	&	4968	&	131	&	259	&	16.94	\\	\hline
W5	&	DN	&	47	&	4968	&	131	&	318	&	14.71	\\	\hline
W6	&	DU	&	65	&	33,264	&	191	&	845	&21.51		\\	\hline
W6	&	DN	&	65	&	33,264	&	191	&	993	&	20.21	\\	\hline

    \end{tabular}
\end{table}

\paragraph{Domains:}

While there exist a set of standard benchmarks for evaluating POMDP algorithms, such as RockSample, or LaserTag \citep{shani2013survey}, these benchmarks do not exhibit significant effort for information gathering. As such domains are the focus of our attention in this paper, we use a set of domains adapted from the contingent planning literature \citep{albore2009translation} that exhibit these properties.

The Wumpus domain (denoted W in the tables below) is an interesting domain where lengthy information gathering sequences of actions are needed, together with multiple sensing actions are different positions. While for each state there is a simple path to the goal, many observations are needed to understand which path applies for the current state. As such, it is a good testing bed for the ability of heuristics to consider sensing actions. A parameter $n$ defines the size of the grid and the number of possible Wumpi, and hence, the amount of needed observations. We create two versions --- one where there is a uniform distribution of possible states, and one where the Wumpi are more likely to be on one side. Actions have deterministic effects.

Although Wumpus requires much information gathering, it does not involve value of information, because sensing actions are mandatory. We hence create a maze domain, where the agent must reach a goal position. In the maze there are several bottlenecks where the agent must pass through one of two cells, one easy to traverse (success probability of 1.0) and the other not. If the cell is hard, or if the agent does not known whether the cell is easy, it can still move through it, but the success probability drops to 0.1. For each possible state there is a path that consists only of easy cells. There are a few cells from which the agent can observe to see whether some cells are easy. The agent must balance between the cost of visiting these observation cells, and moving towards the goal.

Localize (denoted L) is a domain where the agent must navigate a simple maze, with uncertainty about its position, observing nearby walls. We add non uniform stochastic transitions, so the agent should consider the expected cost to the goal. That is, some paths have a higher probability of success than others. Localize domains are similar to the well known Hallway problems \citep{Littman} which are still relatively challenging for modern point-based solvers due to high uncertainty.

In the Colorballs (denoted CB) domain the agent must search colored balls in a small grid and move them to designated positions. The Elog domain (denoted EL) is a standard logistics problem. In these two domains we have uniform initial states and deterministic actions.

Blocks (denoted B) is a standard Blocksworld problem, with uncertainty about the positions of blocks, and stochastic outcomes where blocks may fall on the table. In Doors an agent must move through a grid, establishing which doors are open. We also add a version where doors can be opened. Some doors are easier to open and some more difficult, implemented using stochastic effects. For some doors the agent cannot know whether they are easy or not to open, but can deduce it from succeeding or failing to open a closed door.

Table~\ref{tbl:domains} shows the properties of the domains. There are several types of some problems, exhibiting either deterministic or stochastic transitions (denoted D or S under Type), and either uniform or non-uniform initial belief (denoted U or N). 
For problem size we report the number of facts ($|P$) in the problem description, without facts whose value is fixed, such as the representation of adjacent cells in a grid. Potentially, there can be $2^{|P|}$ possible states, and a naive representation of the belief state would be unmanageable. However, but due to dependencies between the facts the actual number of states is much lower. For example, in Blocks, block $A$ cannot be at the same time on block $B$ and on block $C$. Also, we can never reach a state where block $A$ is on block $B$ and block $B$ is on block $A$. Hence, we explicitly compute the number of possible states ($|S|$) in each of the problems. As can be seen, this number is substantially lower, allowing us to represent the belief state efficiently.

The number of actions ($|A|$) may be misleading, because while in some domains, such as Blocks, there are many actions, typically only for a few of them the precondition hold at a given belief state. $|B|$ is the amount of belief states that were visited by any of the methods while acting under the final policy. RTDP-BEL must visit at least these beliefs, and their neighbors, before converging to an optimal policy. $E[C]$ is the average cost of the best policy. Methods that did not reach this expected cost within the time limit were considered as fail to converge.

\begin{table}[t]
\caption{Comparing the time (seconds) until convergence of delete relaxation heuristics. For both state-based and belief-based we compare $\hmax,\hadd,\hff$.}
\label{tbl:relaxations}
\centering
    \begin{tabular}{|l|c||c|c|c||c|c|c|}
    \hline
    \multicolumn{2}{|c||}{Problem}&\multicolumn{3}{|c||}{Belief-based}&\multicolumn{3}{c|}{State-based} \\ \hline
        Name&	Type	&Max	&Add	&FF	&Max	&Add	&FF	\\ \hline\hline
L5&	DU&	0.42&	0.43&	0.22&	\textbf{0.03}&	\textbf{0.03}&	\textbf{0.03}	\\ \hline
L5&	SU&	1.44&	1.47&	1.23&	0.2&	\textbf{0.18}&	0.29	\\ \hline
L7&	DU&	3.04&	3.08&	1.38&	\textbf{0.11}&	0.12&	0.12	\\ \hline
L7&	SU&	22.8&	21.6&	8.3&	2.89&	\textbf{2.54}&	9.27	\\ \hline
L9&	DU&	16.1&	15.3&	10.1&	0.28&	\textbf{0.27}&	0.29 
   \\ \hline
L9&	SU&74.4&	72.9&	56.4&	32.1&	\textbf{31.8}&	64.4
				\\ \hline	

EL&	DU&	15.7&	8.07&	11.8&	14&	\textbf{6.88}&	12.4\\ \hline
CB&	DU&	1.72&	\textbf{1.18}&	1.27&	1.72&	1.65&	1.69\\ \hline

B7&	DU&	0.97&	0.24&	\textbf{0.04}&	1.18&	0.42&	0.27\\ \hline
B7&	DN&	1.15&	0.33&	\textbf{0.03}&	1.25&	0.49&	0.3\\ \hline
B7&	SU&	231&	288.9&	\textbf{24.2}&	193.6&	291.8&	210.3\\ \hline

D7&	DU&	22.4&	20.7&	15.8&	9.84&	9.88&	\textbf{9.5}\\ \hline
D7&	DN&	8.27&	8.38&	\textbf{4.2}&	5.69&	5.38&	5.48\\ \hline
D7&	SN&	0.05&	0.05&	\textbf{0.02}&	0.74&	0.75&	0.74\\ \hline

M7,1&	DU&	23.4&	22.2&	\textbf{2.44}&	8.05&	8.2&	7.99\\ \hline
M7,2&	DU&	99.1&	62.5&	\textbf{3.79}&	27.7&	31&	31.1\\ \hline
W4&	DU&	0.22&	0.22&	\textbf{0.16}&	0.21&	0.2&	0.2\\ \hline
W4&	DN&	0.16&	0.15&	\textbf{0.07}&	0.21&	0.2&	0.2\\ \hline
W5&	DU&	6.8&	7.05&	\textbf{4.92}&	8.57&	8.58&	7.83\\ \hline
W5&	DN&	6.4&	6.64&	\textbf{3.3}&	6.02&	6.13&	6.16\\ \hline
W6&	DU&	152.4&	127.1&	\textbf{37.2}&	150.7&	133.2&	133
	\\ \hline
W6&	DN&	146.7&	124.3&	\textbf{25.2}&	135.3&	142.3&	131.7\\ \hline

    \end{tabular}
\end{table}

\begin{table}[t]
\caption{Comparing the time (seconds) of $\hff$ to MDP heuristics.}
\label{tbl:runtime}
\centering
    \begin{tabular}{|l|c||c|c|c|c|c|}
    \hline

        Name&	Type	&$\hff(b)$	&$\hff(s)$	&$\qmdp$	&ML	&Flat	\\ \hline\hline

L5	&	DU	&	0.22&		0.03	&	0.01	&	\textbf{0.003}	&	0.02	\\	\hline
L5	&	SU	&	1.23&		0.28	&	0.03	&	\textbf{0.02}	&	0.27	\\	\hline
L7	&	DU	&	1.38&		0.12	&	0.05	&	\textbf{0.01}	&	0.11	\\	\hline
L7	&	SU	&	8.3&		13.5	&	0.23	&	\textbf{0.15}	&	13.5	\\	\hline
L9	&	DU	&	10.1&		0.32	&	0.12	&	\textbf{0.01}	&	0.29	\\	\hline
L9	&	SU	&	83.3	&	89.2	&	\textbf{0.19}	&	0.27	&	114.3	\\	\hline
														
EL	&	DU	&	12.1	&	17	&	\textbf{6.02}	&	\textbf{5.56}	&	18.1	\\	\hline
														
CB	&	DU	&	1.25	&	1.47	&	1.98	&	1.45	&	\textbf{1.0}	\\	\hline
														
B7	&	DU	&	\textbf{0.04}	&	0.26	&	1.51	&	0.73	&	7.18	\\	\hline
B7	&	DN	&	\textbf{0.03}	&	0.31	&	1.78	&	1.42	&	7.53	\\	\hline
B7	&	SU	&	\textbf{24.2}	&	210.3	&	$\times$	&	$\times$	&	$\times$	\\	\hline
														
D7	&	DU	&	15.8&		\textbf{9.95}	&	30.9	&	16.9	&	27.3	\\	\hline
D7	&	DN	&	\textbf{4.2}&		5.32	&	25.4	&	9.71	&	28.2	\\	\hline
D7	&	SN	&	\textbf{0.02}&		0.92	&	12.3	&	0.4	&	$\times$	\\	\hline
														
M7,1	&	DU	&	\textbf{2.44}&		8.24	&	10.3	&	8.08	&	23.8	\\	\hline
M7,2	&	DU	&	\textbf{3.79}&		36.1	&	53.7	&	35.2	&	82.6	\\	\hline
														
W4	&	DU	&	\textbf{0.16}&		0.22	&	0.3	&	0.15	&	0.45	\\	\hline
W4	&	DN	&	\textbf{0.07}&		0.21	&	0.27	&	0.17	&	0.43	\\	\hline
W5	&	DU	&	\textbf{4.92}&		7.26	&	9.17	&	5.16	&	57.6	\\	\hline
W5	&	DN	&	\textbf{3.3}&		5.22	&	7.19	&	4.44	&	51.6	\\	\hline
W6	&	DU	&	\textbf{37.2}&		260.7	&	246.9	&	128.5	&	$\times$	\\	\hline
W6	&	DN	&	\textbf{25.2}&		239.1	&	219.5	&	129.8	&	$\times$	\\	\hline

    \end{tabular}
\end{table}

\begin{table}[t]
\caption{Comparing the number of RTDP iterations until convergence of $\hff$ to MDP heuristics.}
\label{tbl:iterations}
\centering
    \begin{tabular}{|l|c||c|c|c|c|c|}
    \hline

        Name&	Type	&$\hff(b)$	&$\hff(s)$	&$\qmdp$	&ML	&Flat	\\ \hline\hline

L5	&	DU	&	106.2	&	141.2	&	141.8	&	\textbf{44}	&	136	\\	\hline
L5	&	SU	&	384	&	770	&	162.6	&	\textbf{135.}2	&	886	\\	\hline
L7	&	DU	&	188	&	244.6	&	208.8	&	\textbf{52.8}	&	250.6	\\	\hline
L7	&	SU	&	1942	&	18780	&	\textbf{564}	&	644.2	&	23560	\\	\hline
L9	&	DU	&	287.2	&	418	&	256.2	&	\textbf{56.4}	&	436	\\	\hline
L9	&	SU	&	15800	&	91400	&	\textbf{401}	&	977	&	137800	\\	\hline
														
EL	&	DU	&	4610	&	7370	&	\textbf{2800}	&	\textbf{2776}	&	7990	\\	\hline
														
CB	&	DU	&	\textbf{485}	&	1662	&	1624	&	1554	&	2224	\\	\hline
														
B7	&	DU	&	\textbf{12.0}	&	120.6	&	99.2	&	60.8	&	3999	\\	\hline
B7	&	DN	&	\textbf{17.4}	&	229.4	&	216.4	&	151.8	&	9308	\\	\hline
B7	&	SU	&	\textbf{4782}	&	$\times$	&	$\times$	&	$\times$	&	$\times$	\\	\hline
														
D7	&	DU	&	\textbf{3456}	&	6640	&	5948	&	4612	&	18060	\\	\hline
D7	&	DN	&	\textbf{1109}	&	2904	&	2792	&	1672	&	45180	\\	\hline
D7	&	SN	&	\textbf{8.8}	&	38.8	&	78.4	&	54.4	&	$\times$	\\	\hline
														
M7,1	&	DU	&	\textbf{141.4}	&	2854	&	1818	&	2004	&	6440	\\	\hline
M7,2	&	DU	&	\textbf{234.6}	&	12460	&	10680	&	14560	&	13520	\\	\hline
														
W4	&	DU	&	\textbf{45.6}	&	220.8	&	201.4	&	136.8	&	706	\\	\hline
W4	&	DN	&	\textbf{93.6}	&	215.4	&	179.6	&	190.6	&	575	\\	\hline
W5	&	DU	&	\textbf{600}	&	3406	&	3416	&	2262	&	11920	\\	\hline
W5	&	DN	&	\textbf{858}	&	2140	&	2100	&	1868	&	18640	\\	\hline
W6	&	DU	&	\textbf{2012}	&	10830	&	10910	&	8280	&	$\times$	\\	\hline
W6	&	DN	&	\textbf{2326}	&	32020	&	29080	&	25040	&	$\times$	\\	\hline

    \end{tabular}
\end{table}

\subsection{Results}

Table~\ref{tbl:relaxations} shows the average runtime until convergence for our delete relaxation heuristics. For both state-based (Section~\ref{scn:StateHff}) and belief-based (Section~\ref{scn:BeliefHff}) we compare the 3 alternatives --- $\hmax,\hadd,\hff$. For Localize (L) problems, where all information gathering is myopic, the state-based methods are best. State-based FF is less successful, especially in the stochastic case.
For the rest of the problems, the belief-based FF heuristic using our belief-based method is better on many domains, but truly excels on the more difficult problems --- the hardest version of Blocks, with stochastic transitions, the Maze problems, that require much information gathering, and Wumpus. As such, we continue our empirical estimation focusing on $\hff$.

Table~\ref{tbl:runtime} compares the runtime of $\hff$ to MDP based methods and a flat heuristic, that assigns the same heuristic value to all states. Table~\ref{tbl:iterations} shows the number of RTDP iterations until convergence.

Localize (L) problems have relatively small state and action spaces. The agent is unclear about its current state, but can gather information without investing much cost. Observations are necessary to achieve preconditions of actions, such as knowing that there is no wall before moving in a particular direction, and hence must be activated in the RTDP-BEL trajectories. Thus, in these problems the MDP-based techniques excel. The most likely state heuristic is best in the deterministic version, attesting to the relative simplicity of these POMDPs. Logistics (EL) has similar properties.

We report the number of iterations as there is typically a tradeoff between the ability to produce accurate heuristic estimates, and the time it takes to compute these estimations. As can be seen, in localize,  $\hff(b)$ often requires less iterations, but it is still much slower, because in this domain the MDP-based methods only need to take the values from the MDP value function, requiring almost no computation time.

In color blocks (CB), one must use many sensing actions, but always myopically, with no need to invest any effort in gathering information. In this domain, the flat heuristic is best, because any heuristic effort is wasted.

The blocks domain (B) is more difficult. It has a huge number of actions, far more than any other domain, and also a large state space. In this domain, the $\hff(b)$ and $\hff(s)$ methods, that allow RTDP-BEL to avoid exploring many irrelevant actions, work better than the the MDP-based methods. For the stochastic version, only $\hff(b)$ converged in time.

In doors (D) the MDP-based techniques are inferior. The stochastic version of doors has the largest state space, but the vast majority of states need not be explored, and indeed, all heuristics (Flat excluded) manage to avoid this needless exploration. Here, $\hff(b)$ is fastest.

The true value of our belief-based method is revealed in domains that require significant information gathering effort --- maze (M) and wumpus (W). In these domains, $\hff(b)$, that manages to identify the need for information gathering rapidly, is an order of magnitude better than all other methods, in the larger instances. 
The advantage is even more pronounced when observing the amount of needed iterations until convergence. In domains such as maze, methods that do not reason about the need for information realize the value of sensing only after investigating all other options. In the larger maze problem, e.g., all methods require more than 10,000 iterations before they explore the information gathering sequences, while $\hff(b)$ converges after less than 150 iterations.

We also tried to run a popular point-based method, SARSOP \citep{kurniawati2008sarsop} on these problems. As SARSOP does not support preconditions, we penalize executing an action is a state where the preconditions do not hold by -10,000. SARSOP also implicitly assumes that the goal state is observable. We hence add a ``declare-goal'' action that moves the agent to a terminal state, with a -10,000 penalty if called from a non-goal state. 
Unfortunately, SARSOP was unable to even load the larger problems here. For localize 9 (L9), SARSOP took 7.1 seconds to converge to the same average cost. For doors 7 (D7), SARSOP was unable to load the problem, and on the smaller doors 5 problem, SARSOP needed 6.94 seconds to converge. On wumpus 5 (W5) SARSOP took 12.11 minutes to converge, and on  maze (M7,1) it needed 2.6 hours (not including the 2 hours it took to load the model). 

This is not a completely fair comparison. It might be that implementing a point-based algorithm using the structured representation that we use, with action preconditions, may be competitive with RTDP-BEL. We leave such investigations for future research.

To summarize, in domains where sensing is immediate and necessary, although our $\hff(b)$ method is useful, it does not present a significant advantage, if any, over the standard MDP-based methods. However, in domains where information gathering is critical, which are arguably most appropriate for a POMDP formalization, our method provides an order of magnitude improvement over MDP-based methods, that are incapable of assessing the value of information generated by long sequences of actions.

In most complex domains our method is better in terms of the amount of required iterations, while the MDP-based methods require less iterations on the simpler Localize problems. While the number of iterations is often less important in model-based POMDP applications, it can be very important in reinforcement learning (RL) applications. In such applications one must often interact with the environment, or a costly simulator, to run trajectories in the state space. Reducing the amount of such interactions (or samples), is often more important than reducing the internal computation time. This is known as the sample complexity of an algorithm \cite[e.g.]{kumar2023sample}. It might be that our heuristics can be even more important in RL, in cases where a model can be specified, as we require, but the transition function is unknown. We leave this to future research.

\section{Conclusion}

In this paper we suggested a new heuristic for estimating the value of a belief state in a POMDP, formalized as stochastic partially observable contingent planning. Our heuristic is able to estimate the cost of lengthy information gathering action sequences, and thus can direct the planner to consider such sequences. While this computation requires more time than standard heuristics, such as the popular $\qmdp$ heuristic, we show that in tasks that require significant effort for information gathering, it is worthwhile.

For future research we will investigate using our method in other scenarios, such as in point-based POMDP methods, as well as in the context of reinforcement learning.

\section*{Acknowledgments}

This paper is partially supported by the ISF fund, grant no. 964/22, and by the Helmsley Charitable Trust through the ABC fund.

\bibliographystyle{named}
\bibliography{ijcai24}

\end{document}